\documentclass[]{article}
\usepackage{proceed2e}
\usepackage{graphicx}
\usepackage{natbib}
\usepackage{hyperref}
\usepackage{url}
\usepackage{amsthm}
\usepackage{amsfonts}
\usepackage[dvipsnames]{xcolor}
\usepackage{lipsum}  
\usepackage{multicol}
\usepackage{tikz}
\usepackage{enumitem}

\usetikzlibrary{bayesnet,shapes,arrows}

\usepackage{multicol}
\usepackage{tikz}
\usepackage{lipsum}
\usepackage{array}

\usepackage{amsmath}
\usepackage{amssymb}
\usepackage{mathrsfs}
\usepackage{graphicx}
\usepackage{subfig}
\usepackage{floatrow}
\usepackage{algorithm}
\usepackage{algorithmic}
\usepackage{float}

\newcommand{\Xb}{\boldsymbol{X}}

\newcommand{\xb}{\boldsymbol{x}}
\newcommand{\yb}{\boldsymbol{y}}

\usepackage{times}

\title{Improving Training on Noisy Stuctured Labels}

\author{} % LEAVE BLANK FOR ORIGINAL SUBMISSION.
\author{ {\bf Abubakar Abid} \\
Electrical Engineering Dept. \\
Stanford University\\
Stanford, CA 94305 \\
\And
{\bf James Zou\thanks{Corresponding author: jamesz@stanford.edu.}}  \\
Biomedical Data Science Dept. \\
Stanford University\\
Stanford, CA 94305 \\
}

\begin{document}

\maketitle

\begin{abstract}
 Fine-grained annotations---e.g. dense image labels, image segmentation and text tagging---are useful in many ML applications but they are labor-intensive to generate. Moreover there are often systematic, structured errors in these fine-grained annotations. For example, a car might be entirely unannotated in the image, or the boundary between a car and street might only be coarsely annotated. Standard ML training on data with such structured errors produces models with biases and poor performance. In this work, we propose a novel framework of Error-Correcting Networks (ECN) to address the challenge of learning in the presence structured error in fine-grained annotations. Given a large noisy dataset with commonly occurring structured errors, and a much smaller dataset with more accurate annotations, ECN is able to substantially improve the prediction of fine-grained annotations compared to standard approaches for training on noisy data. It does so by learning to leverage the structures in the annotations and in the noisy labels. Systematic experiments on image segmentation and text tagging demonstrate the strong performance of ECN in improving training on noisy structured labels.  

\end{abstract}

\section{INTRODUCTION}
% \james{Cite refs in intro.}
The quality of labeled data plays a significant role in the performance of supervised machine learning methods trained on the data (\citep{nettleton2010study, hendrycks2019benchmarking}). However, in many settings, it may be difficult to obtain high-quality data, such as due to limited time, budget, or expertise dedicated to the annotation process. This is particularly the case for \textit{fine-grained annotations} (\citep{heller2018imperfect}), which are labels that are applied to individual elements of each input data and they often follow certain structures (we will use the terms \emph{label} and \emph{annotation} interchangeably in this paper). 

For examples, in computer vision, semantic segmentation models are trained on image data in which each pixel is labeled for a class (e.g. car, street) \citep{long2015fully}. Such pixel-level labels are not independent, and systematic errors may be present in the training set and thereby learned by a supervised machine learning algorithm. In natural language processing, the analogous task of name-entity recognition can be seen as operating on fine-grained structured data, in which word or token is labeled with the entity that it represents \citep{lample2016neural}.  

\begin{figure*}[th]
\includegraphics[width=\linewidth]{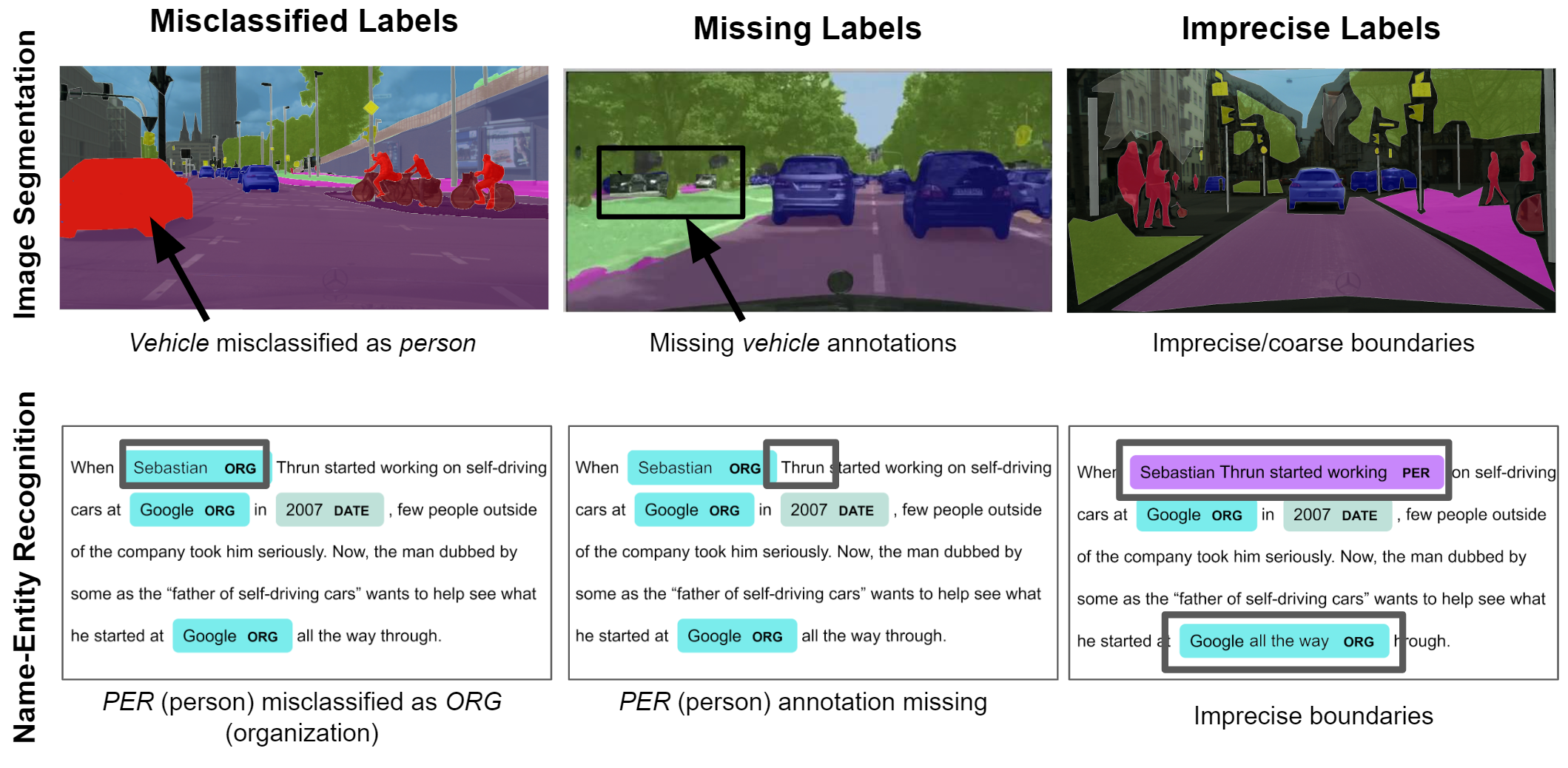}
\caption{\textbf{Typical labeling errors in structured data.} Here, we show examples of errors that commonly occur in two machine learning tasks that operate with structured data: image segmentation (top row) and name-entity recognition (bottom row). Generating precise labels for each pixel or word is very labor intensive. In many images/text, parts of the data is misannotated to be the wrong class (left), or annotations are missing (center), or the borders of the annotations are coarse, e.g. the annotated street doesn't reach all the way to the sidewalk (right). Error-correction Network (ECN) is a general framework to address these structured mistakes in the labels. \label{fig:labeling_errors}}
\end{figure*}

Because of the level of precision needed to finely annotate such structured datasets, it is very common in practice to have datasets with substantial annotation mistakes. This is the case both in widely used public datasets, but even more so in private datasets that are collected and annotated using customized processes. 
In Fig. \ref{fig:labeling_errors}, we provide image and text examples of how various types of complex errors may appear in fine-grain labels. Certain elements could be mislabeled to be a wrong class; entire elements (e.g. a car) could be missing an annotation; often times, the boundaries between different annotated classes (e.g. where does the street begin and sidewalk end) are imprecise. It is important to note that the errors in the labels have many structures and are not independent. For example, errors tend to locally cluster---if a whole car is missed by the labeler, then all of its pixels are misannotated. These structured errors are challenging for standard ML training. Common approaches for training on noisy data are typically developed in settings where there is a \emph{simple} label per data, as is the case in standard classification and regression. They are not well-suited for fine-grained predictions with structured errors within the label of each data. On the other hand, structures in the label enables us to more easily learn to correct the errors. We leverage this idea in developing the new approach of Error-Correcting Networks (ECN).

Human label errors are difficult to avoid because it is extremely labor intensive (and tedious) to precisely annotate and segment all of the individual elements in an image or text.  
In such settings, it is usually the case that a small amount of samples are known to be high-quality data (either by manual quality assurance checks or by dedicating additional annotation resources). In this paper, we propose the ECN method to leverage a small amount of the high-quality datasets, which we refer to as \textit{gold data}, to improve the quality of the \textit{entire} training dataset. Our method is simple to implement and intuitive, and we demonstrate that, even with a relatively small amount of gold data, we can obtain qualitative and quantitative improvements in both computer vision and natural language applications.

\paragraph{Our contributions}
Structured errors in fine-grained labels is a prevalent challenge. However, previous work on label noise has focused on the setting where the label is simple (e.g. a class or a value). We propose a novel and intuitive algorithm of Error-correction Networks (ECN), which can flexibly correct structured errors across diverse domains. Our experiments demonstrate that ECN can substantially improve performance in fine-grained image segmentation/annotation and in text tagging, which are two important and widely used settings. ECN is computationally as well as data efficient---it works well even when there is only a small number of gold standard labeled samples. To the best of knowledge ECN is the first flexible method that can correct diverse types of structured label errors.

\section{RELATED WORKS}

Previous literature has established that noisy labels degrade the performance of supervised machine learning algorithms \citep{nettleton2010study, hendrycks2019benchmarking}. 
Various methods have been proposed to  identify noisy labels and mitigate their effect on the learning algorithm. Broadly, such methods can be divided into three general approaches. The first approach attempts to identify the incorrect labels, and remove them from the dataset \textit{before} training a learning algorithm. Some of these methods require a small amount of data that is known to be correctly labeled \citep{hendrycks2018using, ghorbani2019data}; other methods do not require any such data, and yet, under certain assumptions, can detect incorrectly-labeled data and remove them from the training dataset \citep{brodley1999identifying, kanj2016editing}. 

The second approach embeds the identification of mislabeled as part of the learning algorithm. These methods modify the learning objective itself to \textit{simultaneously} estimate the noise of each sample and use the sample to train the model. Examples of such methods include the expectation-maximization-based methods \citep{xiao2015learning, goldberger2016training}, which treat true labels as latent variables to be recovered, as well as methods that reweigh each sample based on an estimate of the reliability of that sample \citep{ren2018learning}. The third approach is to ignore the fact that some data may be mislabeled, and train a learning algorithm as one would in the case that the data is fully labeled correctly. This is valid in regimes where the amount of correctly-labeled data is large and the label noise is not systematic \citep{rolnick2017deep}.

The vast majority of previous research considers noisy labels in the setting where the learning algorithm is classification or regression, i.e. in which the label for each sample is a single value, either a class category or number. Comparatively little attention has been paid to fine-grained structured labels, such as image segmentation data. Methods proposed for classification/regression may not be appropriate for structured labels, as the types of errors that occur in structured data are different than those that occur in classification/regression. For example, a given image segmentation may be partially accurate and partially mislabeled. In Fig. \ref{fig:labeling_errors}, we demonstrate three types of errors that are common in and somewhat specific to structured labels. 

The problem of noisy structured labels has been observed in specific domains. For example \citet{heller2018imperfect} considered the problem of imprecise segmentation labels for liver CT scans. However, no general framework has been proposed for addressing the problem of training machine learning models on noisy structured data. In this paper, to the best of our knowledge, we propose the first general algorithm for addressing errors in structured labels, using the framework of error-connecting networks.

\section{METHOD}

%\james{Use consistent terminology: either elements or features; annotation or label; $g$ or $g_j$.}

We are given samples from a potentially corrupted dataset $\{(\tilde{\xb}^{(i)}, \tilde{\yb}^{(i)})\}_{i=1}^n$, where $\tilde{\xb}^{(i)}, \tilde{\yb}^{(i)} \in \mathbb{R}^d$. We also have some gold data $\{(\xb^{(i)}, \yb^{(i)})\}_{i=1}^m$, where ${\xb}^{(i)}, {\yb}^{(i)} \in \mathbb{R}^d$ and generally $m \ll n$. For each label $\tilde{\yb}^{(i)}$ in the corrupted dataset, we have an unobserved true label $\bar{\yb}^{(i)}$. We will assume for convenience in this paper that the input features and labels have the same dimensionality i.e. that each $\tilde{\xb}^{(i)}, \xb^{(i)}, \tilde{\yb}^{(i)}, \bar{\yb}^{(i)}, \yb^{(i)}$ is a vector of length $d$. This is without loss of generality, since if some elements are not annotated, we can just define that they belong to the background class. We will refer to each component of these vectors as an \textit{element}, and allow the elements to be real numbers or tuples of real numbers. For example, in the case of image segmentation, the $j^{\text{th}}$ element of $\xb^{(i)}$, denoted as $\xb^{(i)}_j$ may be the RGB-tuple of the $j^{\text{th}}$ pixel in the image, while $\yb^{(i)}_j$ would be the corresponding one-hot tuple that designates the class label for that pixel. Throughout the paper, the superscript denotes the sample and the subscript denote a particular element in one sample. We may drop the sample index $i$ for notational simplicity.

\begin{figure}[th]

{
  \tikz{ %
        \tikzset{plate caption/.append style={below left=0pt and 0pt of #1.south east}}
        \node[latent, rectangle] (s) {$\bar{\yb}_j$} ; %
        \node[obs, rectangle, right= 1.5cm of s] (z) {$\text{RS}_j(\tilde{\xb}, \tilde{\yb}_{-j})$} ; %
        \node[obs, rectangle, below=of s] (x) {$\tilde{\yb}_j$} ; 
        \path[->]
        (s) edge node[right] {} (x) 
        (z) edge node[right] {} (x);
        % (w) edge node {} (x);
        \plate[inner xsep=1cm, inner ysep=0.35cm, xshift=0cm, yshift=0cm] {plate1} {(x) (z)} {$j=1 \ldots d$}; %

      }}

\caption{\textbf{Modeling Errors in Structured Labels.} We propose a simple model for errors in structured labels. Suppose there are $d$ elements in each data (e.g. $d$ pixels or $d$ words). Each observed element-wise label $\tilde{\yb}_j$ is a probabilistic function of the true unknown element-wise label $\bar{\yb}_j$ and some Relevant Subset of elements $\text{RS}_j$ which is a subset of the input features $\tilde{\xb}$ and remaining observed label $\tilde{\yb}_{-j}$. Observed quantities are shaded.\label{fig:error_model}}
\end{figure}

\begin{figure*}[t!]
\includegraphics[width=\linewidth]{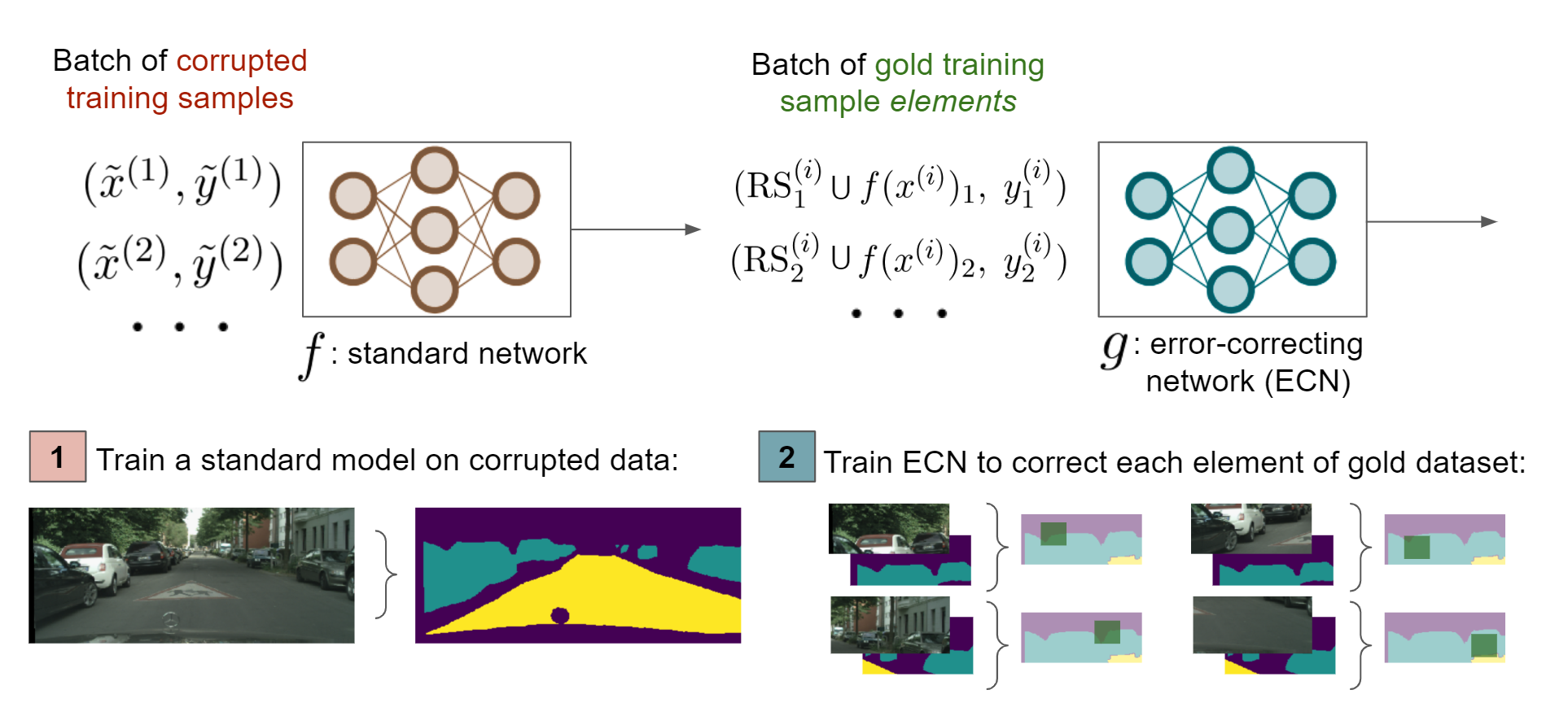}
\caption{\textbf{Overview of the ECN Method.} Our approach to train on corrupted data consists of several steps. First, we train a standard ``base model" $f$ on the corrupted data; $f$ can be any differentiable model. We then use $f$ to make (noisy) label predictions on the gold data. Next, we then train a separate model on \textit{each element of the gold data}, with the goal of predicting the correct label from the noisy label and the relevant subset of features/labels. Here $f(x^{(i)})_j$ denotes the prediction for element $j$ in the $i$-th gold training sample, and $\mbox{RS}_1^{(i)_j} \cup f(x^{(i)})_j$ refers to the concatenation of the predicted element with its relevant subset. The error correction network is $g$, and is shared across all of the elements to improve efficiency. The networks $f$ and $g$ can be trained together end-to-end (see Section \ref{subsection:error-connecting-networks}).   The architecture is shown on the left, and a description with example inputs on the right. \label{fig:overview_method}}
\end{figure*}

\subsection{MODELING STRUCTURED ERRORS IN FINE-GRAINED LABELS}

We first propose a flexible probabilistic model that can generate the kinds of errors that are observed in structured data (and demonstrated Fig. \ref{fig:labeling_errors}). We note that errors in structured data are generally \textit{not} independent across the elements of the sample.
Concretely, consider the case of \textit{misclassified labels} for an image segmentation task. If a particular pixel is misclassified (e.g. \textit{person} instead of \textit{vehicle}), it is likely that neighboring pixels are also misclassified as such. The same holds true for missing labels and imprecise labels. Thus, a probabilistic graph that models the relationship between the true label and the observed labels should take into account a relevant subset of the observed labels (e.g. labels of neighboring pixels). In some cases, a more complete model would also take into account a relevant subset of the input features (e.g. values of  the neighboring pixels), as they may explain the errors in the labels. As a concrete example, it might be the case that lightly-colored objects in an image might be easier to miss during annotation. We will denote the subset of $\tilde{\xb}$ and $\tilde{\yb}_{-j}$\footnote{$\tilde{\yb}_{-j}$ is the vector of all the labels except for the label of the $j$-th element.} that is relevant for the label $\tilde{\yb}_j$ by  $\text{RS}_j$. It could be that $\text{RS}_j = (\tilde{\xb},  \tilde{\yb}_{-j})$; however there could be locality (e.g. between nearby pixels) in many settings, in which case $\text{RS}_j$ would have smaller cardinality and error correction would be easier to learn. See Fig. \ref{fig:error_model} for illustration of probabilistic graphical model

The approach of ECN to use intuition or domain knowledge to identify a reasonably small relevant subset of the sample that, in conjunction with the underlying correct element $\bar{\yb}_j$ allows us to explain the observed sample $\tilde{\yb}_j$. As we will see in Section \ref{subsection:error-connecting-networks}, by choosing an appropriate subset, we leverage structure in annotation errors us to estimate the correct underlying label in an efficient manner.

\subsection{ERROR-CORRECTING NETWORKS}
\label{subsection:error-connecting-networks}

The framework of error-correcting networks consists of two steps. First, we fully train a standard network $f$ on the corrupted dataset $\{(\tilde{\xb}^{(i)}, \tilde{\yb}^{(i)})\}_{i=1}^n$.  Second, we train a much smaller network $g$ to correct mistakes using the correct labels provided for the gold dataset. We refer to $g$ as the \textit{error-correcting network}, as it aims to produce a corrected estimate of each element of the label. Because we do not have corresponding pairs of corrupted and corrected labels, the input to $g$ is instead the prediction $f(\xb)_j$ on the gold data $\xb$ from our standard network $f$, concatenated with the relevant subset of input features and labels $RS_j$. The output of $g$ is the correct label for the element $\yb_j$, which is available for our gold dataset. In principle, there could be a different network $g_j$ to correct the label of each element $j$. For computational and sample efficiency, we use a single network $g$ for all of the elements, and this works well in our experiments. 
The workflow of ECN illustrated schematically in Fig. \ref{fig:overview_method} and pseudocode is provided in Fig. \ref{fig:pseudo}.

In this sense, we utilize and invert the probabilistic model defined previously to estimate the underlying true labels. If the relevant subset that we have defined is too simple (e.g. if it is missing some of the features needed to explain the observed labels), then the error-correcting network $g$ will not be accurate. If it is too complex (e.g. if we simply include the entire input sample), then the complexity of the error-correcting network will need to be correspondingly higher and it will be harder to train.

\begin{figure*}[th]
\begin{algorithm}[H]
  \caption{Training on Noisy Structured Data}
  \label{alg:ecn}
  \begin{algorithmic}
  \STATE {\bfseries Input:} Corrupted dataset $\{(\tilde{\xb}^{(i)}, \tilde{\yb}^{(i)})\}_{i=1}^n$, gold dataset $\{(\xb^{(i)}, \yb^{(i)})\}_{i=1}^m$, in which the dimensionality of the features and labels is $d$, relevant subset selector $\text{RS}$, standard network $f_\theta$, error-correcting network  $g_\phi$. For each network: number of steps $S_1, S_2$, batch size $B_1, B_2$, optimizer, loss
  \vspace{0.1cm}
    \STATE Initialize the parameters of the networks.  
  \FOR {$\text{step} \in \{1 \ldots S_1\}$}
    \STATE Randomly select a batch of corrupted data $\{(\tilde{\xb}^{(i)}, \tilde{\yb}^{(i)})\}$ of size $B_1$ 
    \STATE Update the parameters of $f_{\theta}$ with optimizer and loss on the selected batch
    \ENDFOR
  \vspace{0.1cm}
  \FOR {$\text{step} \in \{1 \ldots S_2\}$}
    \STATE Randomly select a batch of gold data $\{({\xb}^{(i)}, {\yb}^{(i)})\}$ of size $B_2$ 
    \STATE Obtain noisy predictions $\hat{\yb}^{(i)} \equiv f_\theta(\xb^{(i)})$
    \STATE Initialize an empty to list $L$ to hold a batch of samples
  \FOR {$\text{i} \in \{1 \ldots B_2\}$}
  \FOR {$\text{j} \in \{1 \ldots d\}$}
    \STATE Form a sample with input is $\hat{\yb}^{(i)}_j$ concatenated with the relevant subset of feature and label: $RS^{}_j(\xb^{(i)}, \hat{\yb}^{(i)}_{-j})$, and with output is $\yb^{(i)}_j$ 
    \STATE Append this sample to list $L$
    \ENDFOR
    \ENDFOR
    \STATE Update the parameters $\phi$ of the network $g$ based on the optimizer and loss on $L$, the new batch of samples.
    \ENDFOR
  
  \STATE {\bfseries Return:} trained standard network $f_\theta$ and error-correcting network $g_\phi$
  
  \end{algorithmic}
\end{algorithm}
\caption{\textbf{Pseudocode for ECN Framework.} Here, we show the proposed algorithm for training on noisy structured data.\label{fig:pseudo}}
\end{figure*}

 So far, we have described the training of networks $f$ and $g$ as occurring entirely separately in two steps, i.e. we freeze the weights of $f$ during the training of the error-correcting network. However, we may actually train the networks end-to-end during the error-correction step. This would allow us to continue to refine the weights of $f$ while training $g$. This fine-tuning is not necessary for the improved performance of ECN in our experiments, though it could be useful in other applications. Therefore we report the results for the two-step ECN to more clearly demonstrate the power of error correction.
 %For the experiments in this paper, we do not use end-to-end training. 

\begin{table*}[t]
\begin{center}
 \begin{tabular}{| m{2.7cm} | m{1.3cm} |  m{1.4cm} | m{1.4cm} | m{1.4cm} | m{1.4cm} | m{1.3cm} | m{1.3cm} | m{1.3cm}  |}
 \hline
 Dataset  & Clean \textit{(bound)} & Corrupted only & Gold only & Combined & Pseudo-label & ECN \hspace{0.4cm}($\Xb$ only) & ECN \hspace{0.4cm} ($\yb$ only) & ECN (Full) \\ 
 \hline
 \hline
 GMB-Im-Fixed & 0.84 & 0.69 & 0.74 & 0.70 & 0.74 & 0.78 & \textbf{0.80} & 0.78 \\ 
 \hline
 GMB-Im-R & 0.84 & 0.78 & 0.74 & 0.78 & 0.74 & \textbf{0.81} & \textbf{0.81} & \textbf{0.81} \\ 
 \hline
 GMB-Im-V & 0.84 & 0.70 & 0.74 & 0.70 & 0.74 & \textbf{0.79} & \textbf{0.79} & \textbf{0.79}  \\ 
 \hline
 GMB-Im-RV & 0.84 & 0.73 & 0.74 & 0.77 &  0.74 & 0.79 & \textbf{0.80} & \textbf{0.80} \\ 
 \hline
GMB-Mi-Rand & 0.84 & 0.71 & 0.74 & 0.72 &  0.74 & \textbf{0.79} & \textbf{0.79}  & \textbf{0.79} \\ \hline
GMB-Mi-Syst & 0.84 & 0.78 & 0.74 & 0.78 &  0.74 & \textbf{0.81} & \textbf{0.81} & \textbf{0.81} \\ \hline
Cityscapes-Im-100 & 0.79 & 0.70  & 0.52 & 0.69 & 0.49 & 0.69 & 0.63 & \textbf{0.74} \\ 
 \hline
 Cityscapes-Im-250 & 0.79 & 0.70 &  0.63 & 0.72 & 0.63 & 0.69 & 0.62 & \textbf{0.74}  \\  
 \hline
 Cityscapes-Im-500 & 0.79 & 0.70 &  0.60 & 0.71 & 0.59 & 0.70 & 0.68 & \textbf{0.75}  \\  
 \hline
Cityscapes-Mis-50 & 0.79 & 0.69 & 0.57 & \textbf{0.79} & 0.55 & 0.69 & 0.70 & \textbf{0.79}  \\   
 \hline
 Cityscapes-Mis-75 & 0.79 & 0.55 & 0.57 & 0.56 & 0.55 & 0.56 & 0.57 & \textbf{0.76}   \\   
 \hline
\end{tabular}
\caption{\textbf{Performance of ECN and baseline methods.} Here we show the performance of different methods to train models on corrupted data. We use versions of two public datasets, the GMB dataset for NLP tagging and the Cityscapes dataset for image segmentation. See Section \ref{section:experiment} for more information on the datasets and method. 
For the GMB datasets, performance is measured as weighted F1 score, and for the Cityscapes dataset, performance is measured as weighted intersection-over-union (IOU) score. In both cases, higher is better. All reported values are on a hold-out test set.  } \label{tab:1}
\end{center}
\end{table*}

\section{EXPERIMENTAL DETAILS}
\label{section:experiment}

We carry out systematic experiments utilizing two standard public datasets and semi-synthetic modifications of these datasets to measure the efficacy of the error-correcting network framework for the three kinds of errors that we have presented previously (see Fig. \ref{fig:labeling_errors}). We also compare our framework to a variety of baseline algorithms that we describe in this section.

\subsection{DATASETS}

\textbf{Natural language processing} For natural language experiments, we use several semi-synthetic modifications of a standard publicly-available name-entity recognition dataset, the Groningen Meaning Bank (GMB) dataset \citep{Bos2017GMB}. The dataset includes sentences from news articles in which each word has been classified into an associated entity (e.g. a \textit{geographic} entity or an \textit{organization} name). Words that are not named entities are labeled with the background class, \texttt{O}.

The first set of modifications we make change the entity labels in the dataset so that they no longer precisely map to the names in the dataset. This is common type of label mistake especially when the labeler only has time to coarsely indicate where certain element of interest is in the text (Figure \ref{fig:labeling_errors}c). Concretely, the four ``\textbf{im}precise" modifications are as follows:

\begin{itemize}
    \item \textbf{GMB-Im-Fixed}: Each entity label is extended by exactly three words. For example, the sentence ``A court in Poland has fined the magazine publisher." would have the \textit{geographic} entity label applied to the words ``Poland has fined the."
    \item \textbf{GMB-Im-R}: Half of the entity labels are \textbf{r}andomly selected and extended by exactly three words.
    \item \textbf{GMB-Im-V}: Each entity labels is selected and is extended by a \textbf{v}ariable number (between 1 and 3 inclusive) of words. 
    \item \textbf{GMB-Im-RV}: Three-fourths of the entity labels are \textbf{r}andomly selected and are extended by a \textbf{v}ariable number (between 1 and 3 inclusive) of words. 
\end{itemize} 

In addition, we have two versions of the GMB dataset that measure the performance of our framework with \textit{missing} labels. The datasets with ``\textbf{mi}ssing" labels are:

\begin{itemize}
    \item \textbf{GMB-Mi-Rand}: In this modified version of the GMB dataset, 30\% of the entity labels are \textbf{rand}omly dropped. If an entity consists of multiple words (e.g. ``New York City"), then the entire label is dropped.
    \item \textbf{GMB-Mi-Syst}: To simulate more \textbf{syst}ematic errors, we conducted in an experiment in which we removed the entity labels that were missed by an off-the-shelf name-entity recognition library (from the \texttt{spacy} package\footnote{See:  https://spacy.io/usage/linguistic-features}). More precisely, we used the library to identify all the named entities. If an entity was identified by the library and annotated in the original GMB dataset, we kept the original label. If the token was labeled as \texttt{O} by the library, we changed its label to \texttt{O}. If a word was labeled as \texttt{O} in the GMB dataset, we kept the \texttt{O} label. This amounted to changing approximately 15\% of the labels to \texttt{O}.

\end{itemize} 

In each case, the corrupted training dataset is of size 37,407 sentences and the gold dataset of size 960. Performance is measured on a test set of size 9,592 where we know the ground truth annotations. We use standard tokenization methods to convert each word into a dictionary of 19 semantic and syntactic features (see Appendix \ref{appendix:features} for a list of features), and each token is mapped to one of 9 entity categories, including the \texttt{O} entity.

\paragraph{Semantic image segmentation} For image segmentation  experiments, we use the Cityscapes dataset \citep{cordts2016cityscapes}, which includes stereo images taken from vehicles in different cities. Each image has been annotated at one of two resolution levels: in the finely-annotated images, each pixel is precisely mapped into one of 30 classes (we use a subset of 3 classes in our experiments: \textit{vehicle}, \textit{road}, and \textit{other}), while in the coarsely-annotated images is annotated with rough polygons outlining the three different classes.

Here, in the first case, we do not make any modifications to the dataset, but consider how to best leverage the finely-annotated images (which we use as the "gold" images) to improve the coarse labels ("corrupted" labels). We always use 2,975 coarsely-labeled images for training, and 500 finely-annotated images as a holdout test set, but we investigate the effect of having different numbers of an additional gold dataset of finely-annotated images for training. More specifically,

    \textbf{Cityscapes-Im-X} refers to a dataset in which we have \textbf{X} finely-labeled images in our training dataset, in addition to the 2,975 coarsely-labeled images. We will consider $\textbf{X} =$ 100, 250 and 500.

We also consider two semi-synthetic examples in which measure the performance of our framework on \textit{misclassified} data. More specifically, in the \textbf{Cityscapes-Mis-X} datasets, \textbf{X}\% of the images have the \textit{vehicle} category mislabeled as \textit{road}. All of the remaining annotations are correct and finely labeled. We will consider $\textbf{X} =$ 50 and 75. Performance is measured on a holdout test set of size 500 correctly and finely-labeled images.

\subsection{IMPLEMENTATION OF ECN}

We characterize the error-correcting network framework with the following hyperparameters. For GMB datasets, we use a conditional random field (CRF) as the standard base model $f$, implemented using the \texttt{sklearn-crfsuite} Python library. We consider different relevant subsets for each word: only using the word features of the token we are considering ("ECN $x$ only"), only using the noisy predicted labels of the three neighboring ("ECN $y$ only"), or both concatenated ("ECN Full"). 

For the Cityscapes dataset, we used a U-Net architecture for the standard network $f$. We used a standard 2D convolutional network for the error-correcting network, with different relevant subsets for each pixel: only using the pixel values from a 64x64 window surrounding the target pixel ("ECN $x$ only"), only using the labels from a 64x64 window surrounding the target pixel ("ECN $y$ only"), or both concatenated in the channel dimension ("ECN Full"). See more details on all network architectures in Appendix \ref{appendix:architecture}.

When we measured the performance of the ECN networks, we trained \textit{only} on the error-corrected version of the corrupted dataset. We did not also  explicitly train on the gold data. We compared these results to the  performance of several baselines to ensure that our results demonstrated meaningful improvements:

\begin{itemize}
\item\textbf{Corrupted only.} Here, we train a standard base model on only the corrupted data $\{(\tilde{\xb}^{(i)}, \tilde{\yb}^{(i)})\}_{i=1}^n$. This represents the typical performance with no error-correction is carried out. 
\item\textbf{Gold only.} Here, we train a standard base model on the small number of gold data $\{(\xb^{(i)}, \yb^{(i)})\}_{i=1}^m$. 
\item\textbf{Combined.} Here, we merge the corrupted and gold training data and train a model on the total $m+n$ training points.
\item\textbf{Pseudolabel.} For this baseline, we use a base model trained on the gold data to completely \textit{relabel} the corrupted dataset. Previous research has suggested that doing this can provide a boost in performance, and is a regularized alternative to simply merging the datasets together \citep{lee2013pseudo}. 
\item\textbf{Clean.} For a benchmark that represented an upper bound on performance of any method, we trained a segmentation model on 2,975 \textit{finely-annotated} images (image segmentation) or uncorrupted GMB dataset (name-entity recognition). Note that these samples were not available to any of the other algorithms. 
\end{itemize}

\section{RESULTS}

Results from our experiments are summarized in Table \ref{tab:1}. Here, we comment on the results in more depth, and provide typical examples of the corrected labels. 

\subsection{NATURAL LANGUAGE PROCESSING}

\begin{figure}[th]
\includegraphics[width=\linewidth]{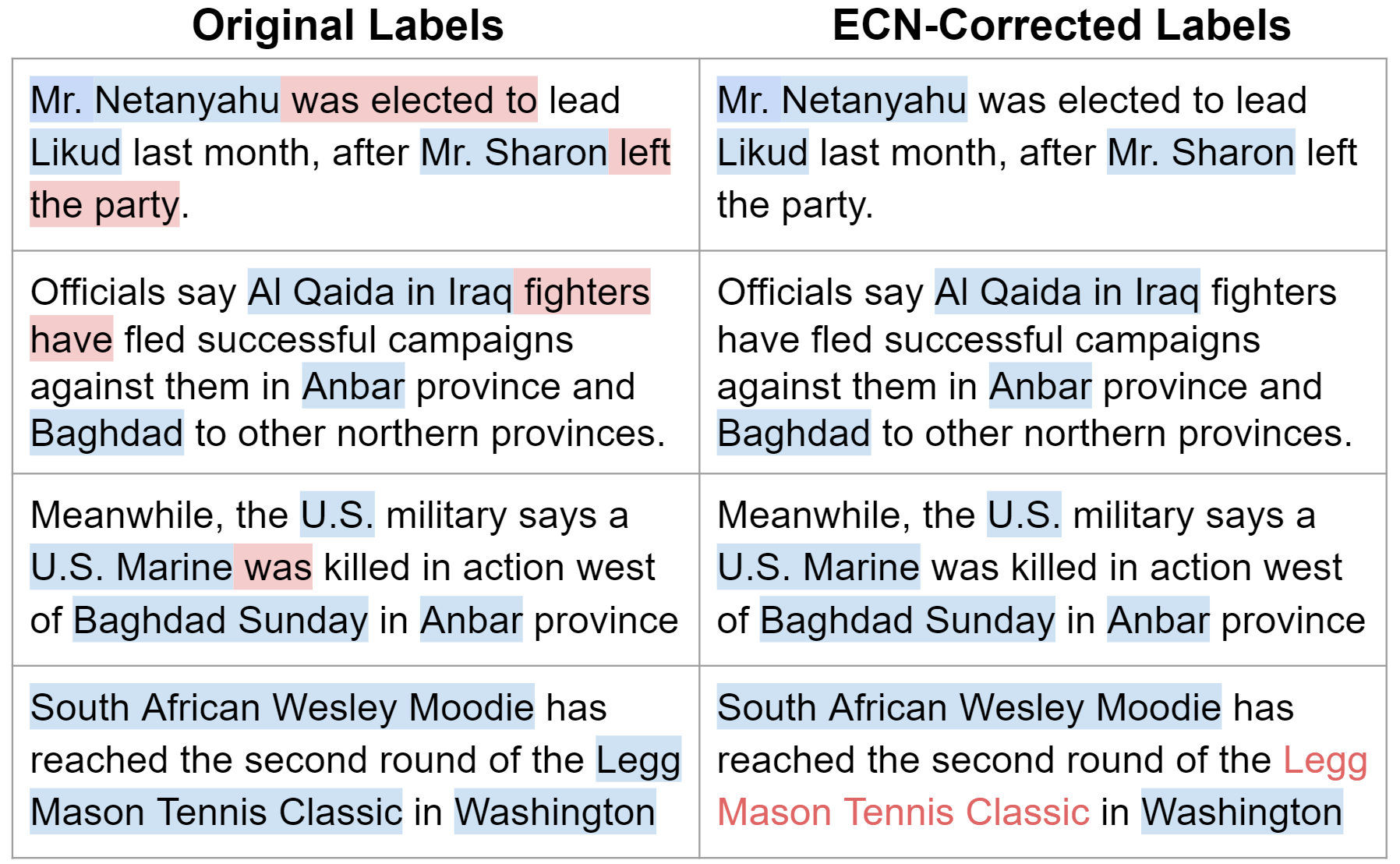}
\caption{\textbf{Example results on GMB Dataset.} Here, we show 4 typical examples from the GMB-RV dataset. The first column shows predictions made from a model trained only on the corrupted dataset. The second column shows predictions after they were corrected by an ECN. The highlights mark all of the words that were marked as a non-O entity. The blue highlights mark words that are correctly labeled, while the red highlights mark incorrect labels. In the bottom row, the ECN-corrected labels mistakenly removes the annotations from ``Legg Mason Tennis Classic" -- this is designated by the red font.  \label{fig:gmb}}
\end{figure}

Across all of the semi-synthetic GMB datasets that we tested, we found that ECNs provided a significant boost in performance compared to the baselines, as measured by the F1 metric macro-averaged  across all of the entities. We found that typically, the ECNs with all of the various relevant subsets that we considered improved on the baselines, with the full ECN generally performing the best. Typically, the next-best approach was to train \textit{only} on the gold data. Pseudolabeling had marginal effect on performance (no difference to the second decimal point). In Fig. \ref{fig:gmb}, we show a few examples of predictions made by a base model trained on the corrupted data versus predictions made by an error-correcting model.

\subsection{SEMANTIC IMAGE SEGMENTATION}

As with the name-entity recognition experiments, we found that ECNs provided a significant boost in performance across all of the Cityscape experiments. Here, the metric we considered was a macro-averaged intersection-over-union (IoU) score. We found that typically, the ECNs with all of the various relevant subsets that we considered improved on the baselines, with the full ECN performing the best. Typically, the next-best approach was to combine the gold and corrupted datasets. In one case (Cityscapes-Mis-50), we found significant improvement (same level as ECN) on the combined dataset, as the extra gold samples allowed the network to correctly disambiguate vehicles from the road. However, in most cases, the improvement in performance on the combined dataset was much smaller. Gold only and pseudolabeling had marginal effect on performance, presumably because the gold dataset was generally too small. 

\begin{figure}[th]
\includegraphics[width=\linewidth]{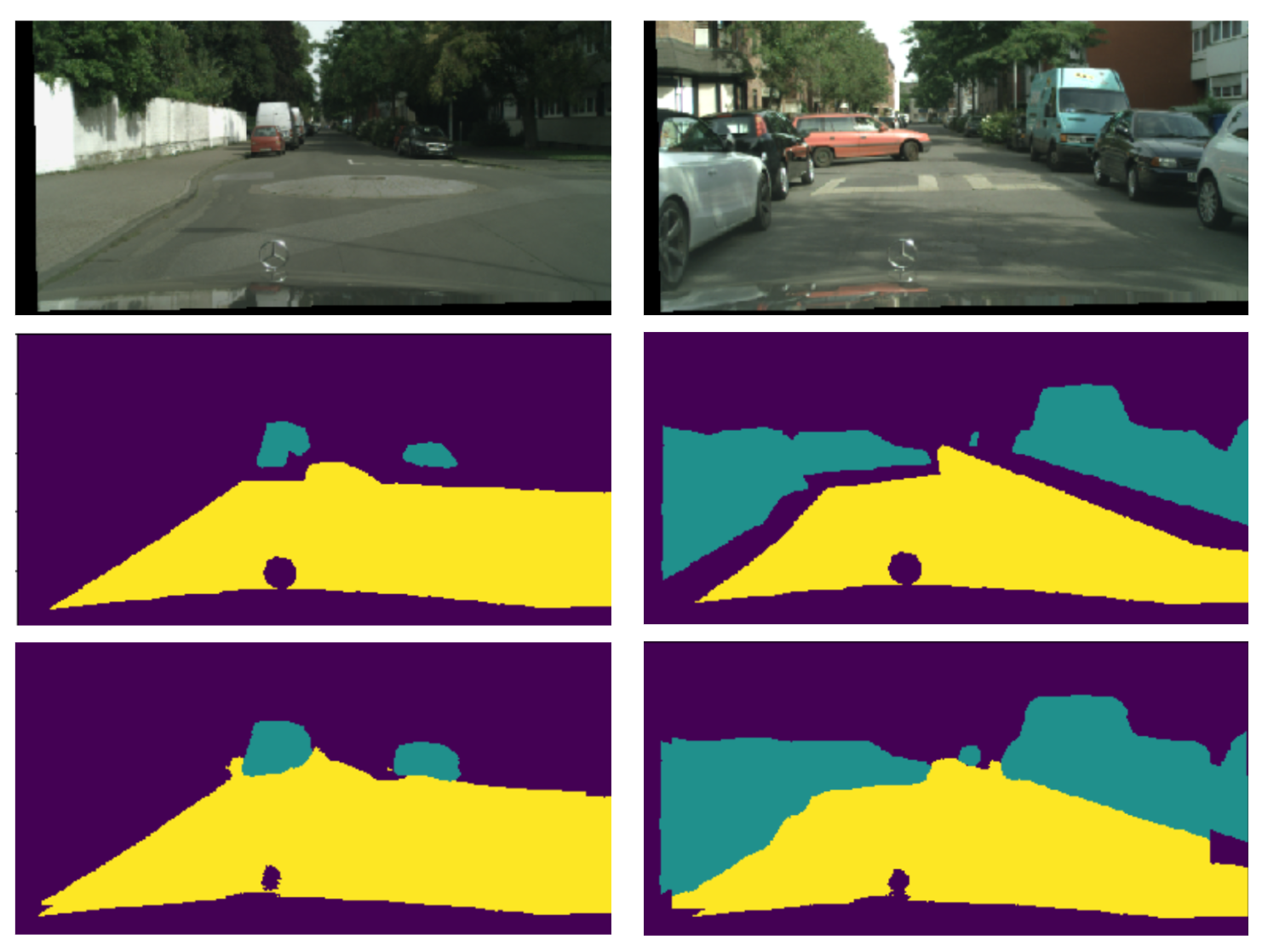}
\caption{\textbf{Example results on Cityscapes-Im-X Dataset.} Here, we show 2 typical examples from the Cityscapes dataset. The first row shows the input camera images that were part of the coarsely-labeled training dataset. The second row shows the examples of the coarse annotations that were provided as part of the dataset. The third row shows the \textit{corrected} labels produced using the ECN trained on 500 gold images. Here, yellow designates the \textit{road} class, green the \textit{vehicle} class, and purple the \textit{background}. \label{fig:cs}}
\end{figure}

In Fig. \ref{fig:cs}, we show a few examples of the original coarsely-labeled annotations versus corrections made by an ECN on the Cityscapes-Im-X dataset. Results on the Cityscapes-Mis-X dataset are in Appendix \ref{appendix:figures-miscl}.

\subsection{SENSITIVITY TO RELEVANT SUBSET}

In Section \ref{subsection:error-connecting-networks}, we mentioned that the size of the relevant subset (RS) has an effect on the resulting performance. Here, we show quantitative results from experiments on the GMB-Fixed dataset. We varied the RS to include varying numbers of neighboring labels (for ``ECN $y$ only"), as well as varying numbers of token features (for ``ECN $X$ only"). We found that, as expected, when no neighboring labels were included, the ECN was unable to correct the observed label. As the number of neighboring labels increased, model performance remained steady, then slightly dipped. When the number of features from $X$ increased, performance generally and gradually increased. The explicit feature set is defined in Appendix \ref{appendix:features}. These results suggest that it is valuable to use domain expertise to craft the Relevant Subsets, though the method is somewhat robust to overly large relevant subsets.

\begin{figure}[th]
\includegraphics[width=\linewidth]{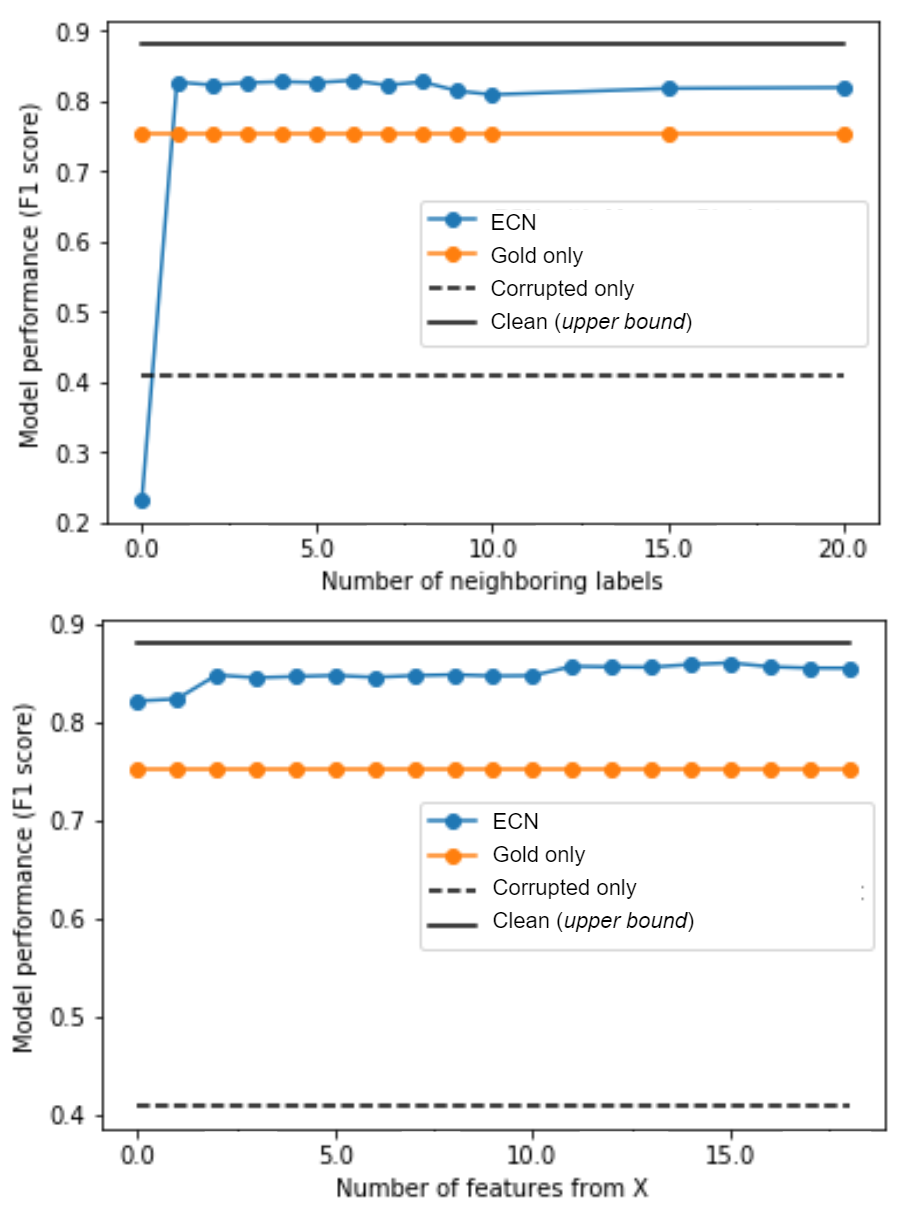}
\caption{\textbf{Effect of Relevant Subset on ECN Performance}. On the left, we plot the performance of the ECN ($y$-only) approach along with several baselines as the number of neighboring labels in the RS is increased. On the right, we plot the performance of the ECN ($x$-only) approach along with several baselines as the number of token features in the RS is increased (added in arbitrary order). Here, we plot the F1 score for one of the tags (GEO), though similar trends were observed across all of the tags in the tagset. \label{fig:sens}}
\end{figure}

\section{DISCUSSION}

In this paper, we have proposed a new framework, the ECN method, to improve the quality of labels in a training dataset of structured labels. The key insight behind our method is it is easier to correct mistakes in structured labels than to make predictions from scratch, especially when the mistakes are structured and localized. Our method is simple to implement and intuitive, and we demonstrate that, even with a relatively small amount of gold data, we can obtain qualitative and quantitative improvements in systematic errors that occur naturally or synthetically in structured data. 

We have tested our method in both computer vision and natural language applications, where we demonstrated that it is able to correct different kinds of structured mistakes that commonly occur in image segmentation labels and name-entity labels respectively. Using the same general framework and architecture, and without changing any significant hyperparameters, we are able to correct misclassified labels, missing labels, and imprecise labels.  We believe that our method is the first general method to designed for noisy structured labels; comparisons to several baseline algorithms demonstrate consistent improvement even when training data is corrupted systematically with imprecise boundaries. In future work, we aim to characterize the ability of this method on other kinds of errors in structured data in further datasets. 

\clearpage
\newpage

\section*{Acknowledgments}

We are grateful to many people for providing helpful suggestions and comments in the preparation of this manuscript. Brainstorming discussions with Ali Abid, Ali Abdalla, and Dawood Khan provided the inspiration and initial formulation of this problem. Feedback from Allen Nie and Bryan He was helpful in guiding the experiments and analyses that were carried out for this paper. J.Z. is supported by NSF CCF 1763191, NIH R21 MD012867-01, NIH P30AG059307, and grants from the Silicon Valley Foundation and the Chan-Zuckerberg Initiative.

\bibliography{main}

\appendix
\newpage
\begin{onecolumn}

\section*{Appendices}

\section{Features in $X$ for CRF in NLP Experiments}
\label{appendix:features}

In the natural language experiments, we used 19 features of each word as the input to the conditional random field models, including in some cases, features of the adjacent words. We list them in Figure \ref{fig:features}, along with an example for reach feature:

\begin{figure}[H]
\includegraphics[width=0.3\linewidth]{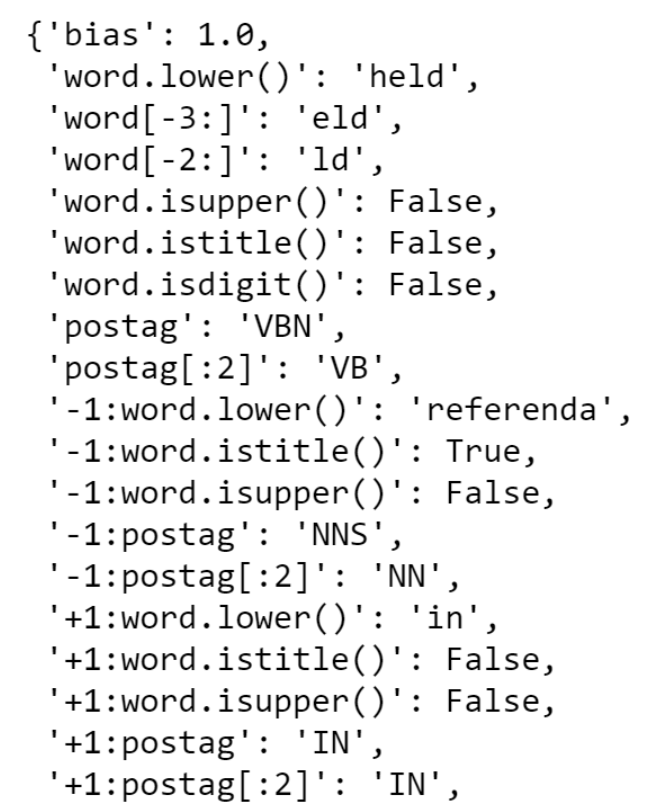}
\caption{Example features from a word used in CRF models. \label{fig:features}}
\end{figure}

\section{Model Architectures}
\label{appendix:architecture}

Here, we explicitly define the architectures used in each step of the algorithm for both sets of experiments.

\subsection{Natural language processing}

For the standard base model, we used the \texttt{sklearn-crfsuite} library's default implementation of a  conditional random field (CRF) trained by gradient descent using the L-BFGS method to predict the entity label for each word. The features constructed from each word used as input to the CRF are listed in Appendix \ref{appendix:features}. Furthermore, we added $L_1$ and $L_2$ regularization to the CRF with regularization coeffcients $c_1$ and $c_2$ both set to 0.1.

For the ECN model, we also used a CRF with the same architecture, except that the input features varied. For the full ECN, we used all of the 19 word-related features listed in \ref{appendix:features}, along with the (corrupted) labels of the 3 neighboring words on both sides of target word. If a word had fewer than 3 neighbors on either side, then we passed in a different `invalid' label for each non-existent neighbor. For ECN ($\Xb$ only), we only passed on the word-related features, and for ECN ($\yb$ only), we only passed in the neighbor labels.

\subsection{Semantic Image Segmentation}

For the standard base model, we used a standard U-net architecture \citep{ronneberger2015u}, with 5 down-sampling blocks and 4 up-sampling blocks, followed by one final convolutional layer. The layers in each block, as defined using the Layers API from the \texttt{keras} library:

\textbf{Downsampling Block 1}
 \begin{verbatim}
    Conv2D(64, (3, 3), padding='same', name='block1_conv1')
    BatchNormalization()
    Activation('relu')
    Conv2D(64, (3, 3), padding='same', name='block1_conv1')
    BatchNormalization()
    Activation('relu')
    MaxPooling2D()
    \end{verbatim}     

\textbf{Downsampling Block 2:} same as Block 1 but with twice as many channels in the convolutional layers.

\textbf{Downsampling Block 3:} same as Block 2 but with three sets of Conv-BatchNorm-Activation, and twice as many channels in the convolutional layers.

\textbf{Downsampling Block 4 and 5:} same as Block 3 but with twice as many channels in the convolutional layers.

Each upsampling block was the transpose of a downsampling block in reverse order along with a concatenation from the output of the corresponding downsampling block. For example:

\textbf{Upsampling Block 1}
 \begin{verbatim}
    Conv2DTranspose(512, (2, 2), strides=(2, 2), padding='same')
    BatchNormalization()
    Activation('relu')
    Concatenate([x, block_4_out])
    Conv2D(512, (3, 3), padding='same')
    BatchNormalization()
    Activation('relu')
    Conv2D(512, (3, 3), padding='same')
    BatchNormalization()
    Activation('relu')
    \end{verbatim}     

\textbf{Upsampling Block 2}: Transpose of Downsampling Block 3

\textbf{Upsampling Block 3}: Transpose of Downsampling Block 2

\textbf{Upsampling Block 4}: Transpose of Downsampling Block 1

For the full ECN model, we used a much smaller convolutional network. The full architecture is defined below. For the ECN ($\Xb$ only) and the ECN ($\yb$ only), we simply replaced the \texttt{inp-y} or the \texttt{inp-x} with random floats respectively:

\begin{verbatim}
    inp_y = Input(shape=(input_window_size, input_window_size, 3))
    inp_x = Input(shape=(input_window_size, input_window_size, 3))
    Concatenate()([inp_x, inp_y])
    Conv2D(8, (4, 4), padding='same', activation='relu')
    MaxPool2D((2, 2), padding='same')
    Conv2D(8, (4, 4), padding='same', activation='relu')
    MaxPool2D((2, 2), padding='same')
    Conv2D(16, (4, 4), padding='same', activation='relu')
    Flatten()
    Dense(40, activation='relu')
    Dense(3, activation='softmax')
    Reshape((1, 1, 3))
\end{verbatim}

\section{Figures for Cityscapes-Mis-X}
\label{appendix:figures-miscl}

Here, we show typical results of the error-correcting networks on \textbf{Cityscapes-Mis-50} and \textbf{Cityscapes-Mis-75} datasets.  In each of the four figures that follow, we show four images:
\begin{itemize}
    \item \textit{top left:} the image used as the input to the semantic segmentation task.
    \item \textit{top right:} the corrupted label in which the \textit{vehicles} have been misclassified as \textit{road}.    
    \item \textit{bottom left:} the output prediction from the base model trained on the corrupted data.
    \item \textit{bottom right:} the output prediction of the error-corrected network (when fed as input the base model prediction, along with relevant subset information) i.e. the corrected label. Note that the elements around the border of the image (64-pixel strip) were not corrected.
\end{itemize}

\begin{figure}[H]
\includegraphics[width=0.8\linewidth]{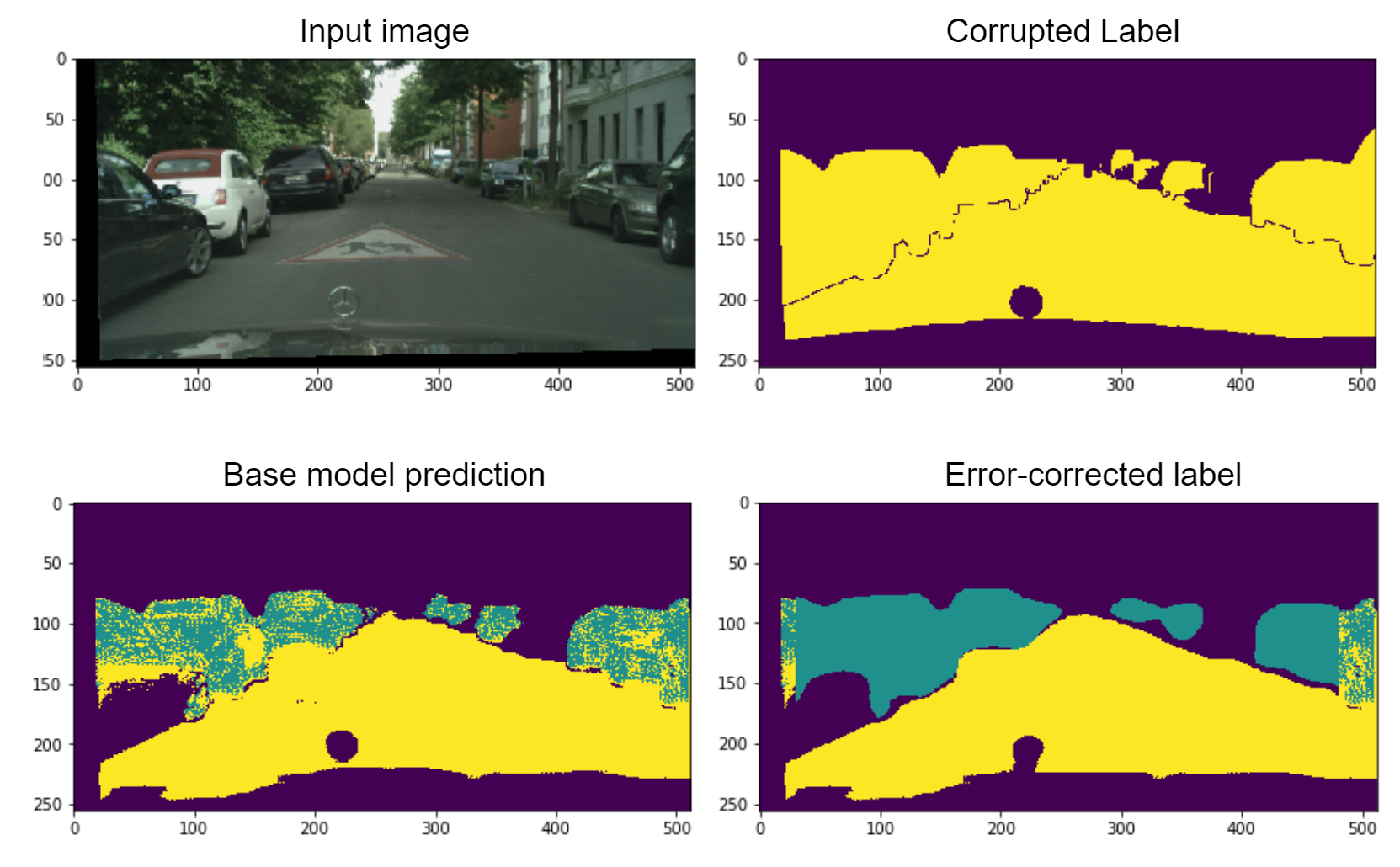}
\caption{Example 1 from Cityscapes-Mis-50 Dataset.}
\end{figure}

\begin{figure}[H]
\includegraphics[width=0.8\linewidth]{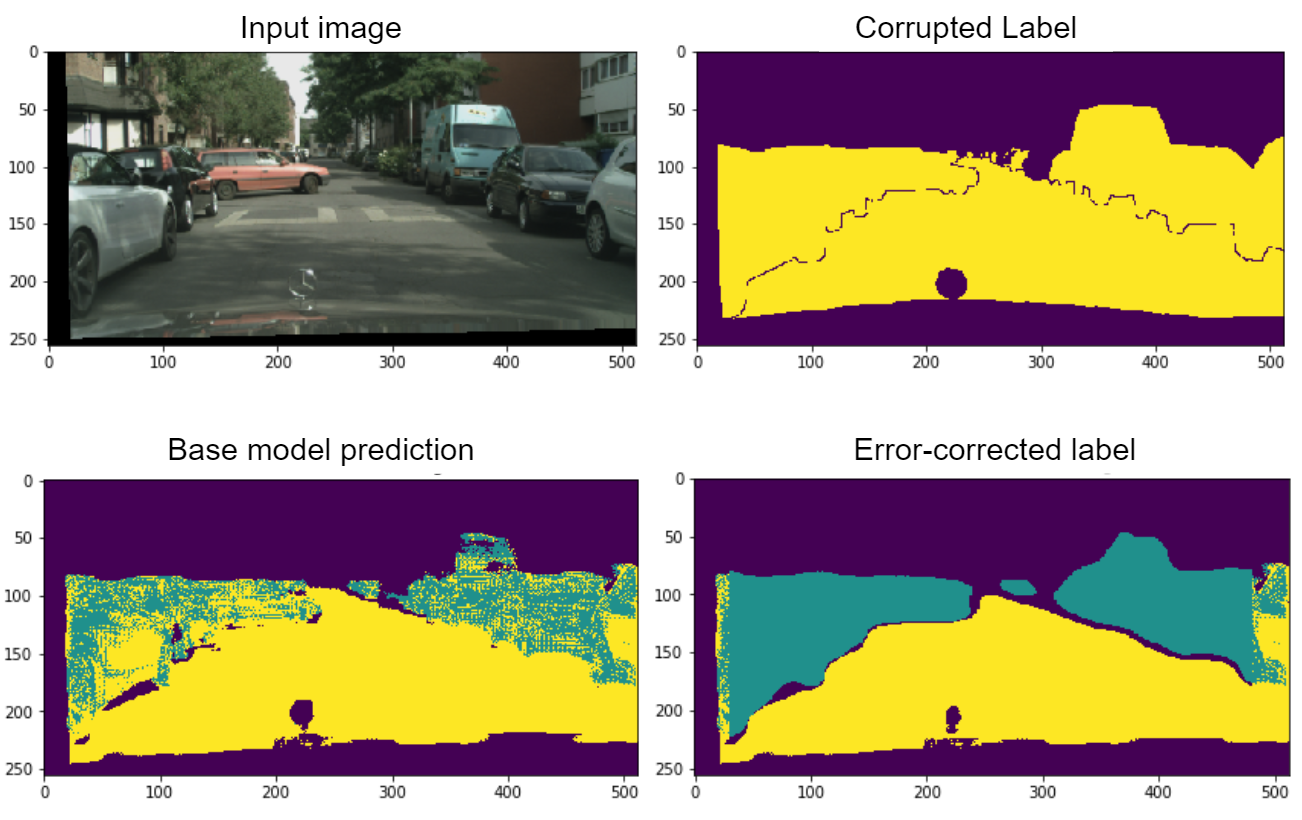}
\caption{Example 2 from Cityscapes-Mis-50 Dataset}
\end{figure}

\begin{figure}[H]
\includegraphics[width=0.8\linewidth]{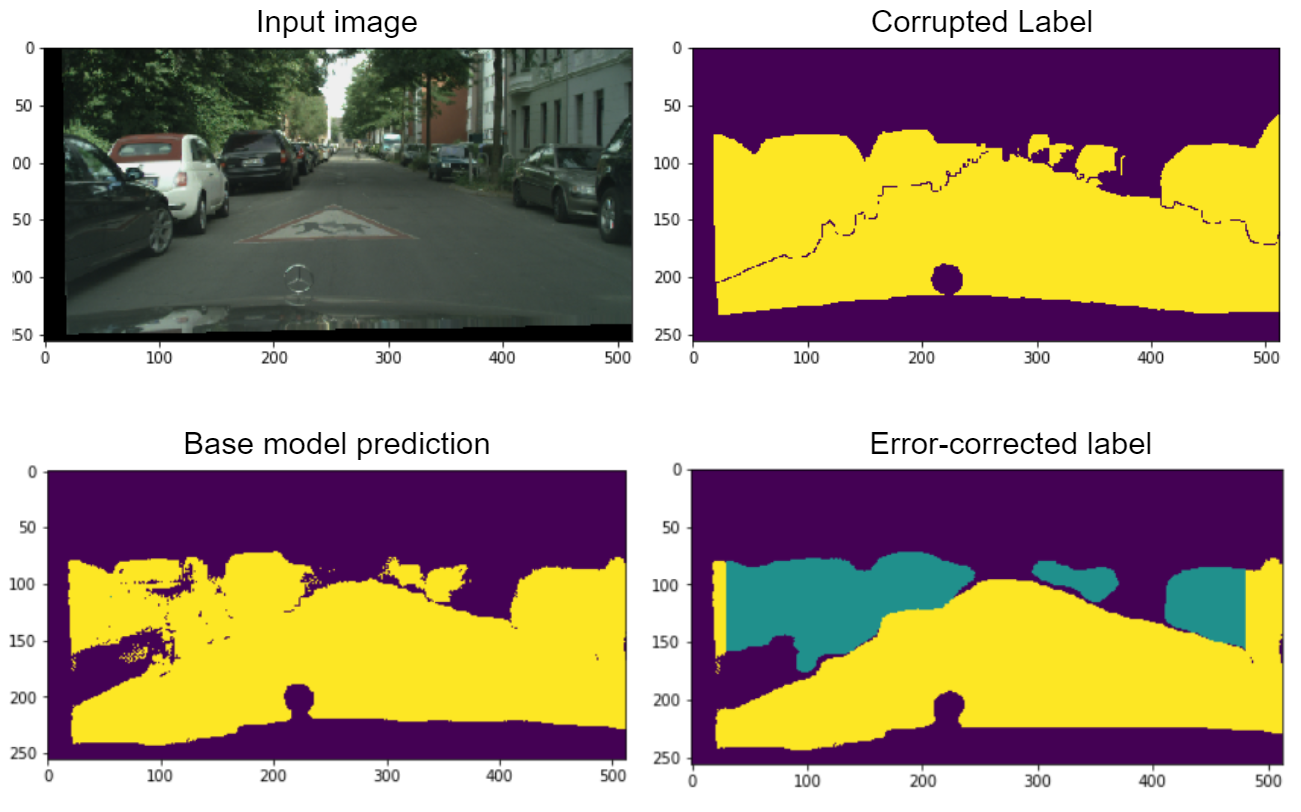}
\caption{Example 1 from Cityscapes-Mis-75 Dataset.}
\end{figure}

\begin{figure}[H]
\includegraphics[width=0.8\linewidth]{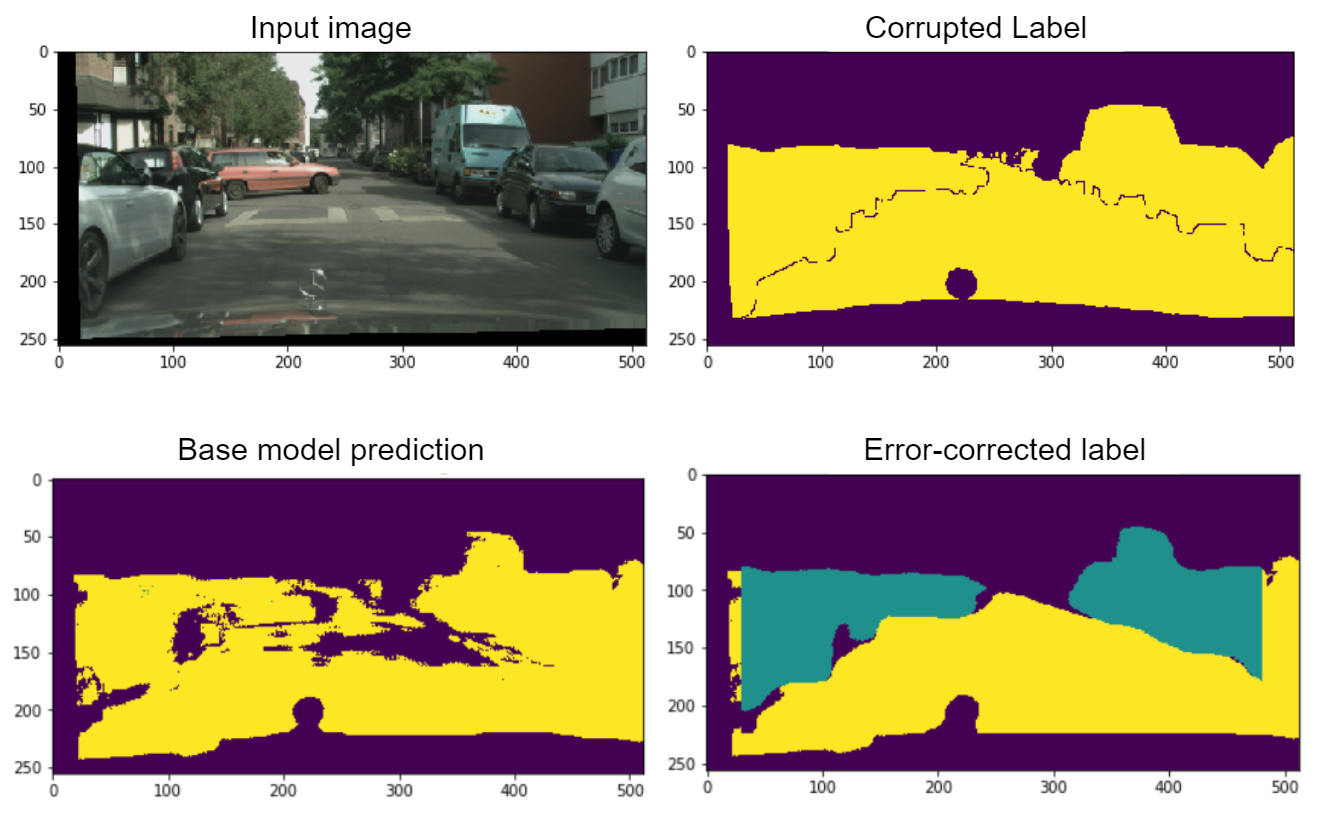}
\caption{Example 2 from Cityscapes-Mis-75 Dataset}
\end{figure}

\clearpage

\end{onecolumn}
\end{document}